\documentclass{article} 

\usepackage[accepted]{icml_r2fm}

\usepackage{amsmath,amsfonts,bm}

\def\eqref#1{equation~\ref{#1}}

\def\1{\bm{1}}

\DeclareMathAlphabet{\mathsfit}{\encodingdefault}{\sfdefault}{m}{sl}
\SetMathAlphabet{\mathsfit}{bold}{\encodingdefault}{\sfdefault}{bx}{n}

\usepackage{microtype}
\usepackage{graphicx}
\usepackage{subfigure}
\usepackage{soul}
\usepackage{booktabs} 
\usepackage{stfloats}

\usepackage{hyperref}
\usepackage[capitalize,noabbrev]{cleveref}
\definecolor{mydarkblue}{rgb}{0,0.08,0.45}
\hypersetup{ %
    colorlinks=true,
    linkcolor=mydarkblue,
    citecolor=mydarkblue,
    filecolor=mydarkblue,
    urlcolor=mydarkblue,
}

\usepackage{amsmath}
\usepackage{amssymb}
\usepackage{mathtools}
\usepackage{amsthm}

\usepackage{tikz}
\icmltitlerunning{Consistency in Language Models}
\begin{document}

\twocolumn[
\icmltitle{Consistency in Language Models:\\
Current Landscape, Challenges, and Future Directions}

\icmlsetsymbol{equal}{*}
\begin{icmlauthorlist} 
\icmlauthor{Jekaterina Novikova}{arva,vg}
\icmlauthor{Carol Anderson}{arva}
\icmlauthor{Borhane Blili-Hamelin}{arva}
\icmlauthor{Domenic Rosati}{du}
\icmlauthor{Subhabrata Majumdar}{arva,vj}
\end{icmlauthorlist}

\icmlaffiliation{arva}{AI Risk and Vulnerability Alliance}
\icmlaffiliation{du}{Dalhousie University}
\icmlaffiliation{vg}{Vanguard, Enteprise AI Research}
\icmlaffiliation{vj}{Vijil}
\icmlcorrespondingauthor{Subhabrata Majumdar}{subho@vijil.ai} 

\vskip 0.3in 
]


\printAffiliationsAndNotice{}  


\begin{abstract}
The hallmark of effective language use lies in consistency: expressing similar meanings in similar contexts and avoiding contradictions. While human communication naturally demonstrates this principle, state-of-the-art language models (LMs) struggle to maintain reliable consistency across task- and domain-specific applications. Here we examine the landscape of consistency research in LMs, analyze current approaches to measure aspects of consistency, and identify critical research gaps. Our findings point to an urgent need for quality benchmarks to measure and interdisciplinary approaches to ensure consistency while preserving utility. 

\end{abstract}



\section{Introduction}
\label{sec:intro}


Consistency---broadly defined as using language similarly in similar settings or avoiding contradictions when using language
---is among the most important forms of generalization 
in the use of language. This ability to maintain consistent outputs is essential for building reliable AI systems that users can trust and depend on. Consistency is both a natural expectation that users have when interacting with language technologies and a prerequisite to deploying them in high-stakes domains \citep{elazar-etal-2021-measuring,jang2022becel,kim2025fostering}. However, most advanced large language models (LLMs) struggle with consistency and frequently demonstrate inconsistent behavior~\citep{elazar-etal-2021-measuring,raj2025improvingconsistencylargelanguage}. Although such examples have been documented in multiple studies, there are no standard approaches to assessing model consistency. As such, there is an ongoing risk of overestimating the performance of state-of-the-art models, as well as of underestimating the risks and potential harms elicited by them.

Despite early attempts to measure and enhance the consistency of language models (LM) and to understand the roots of inconsistency, research on this topic faces multiple challenges. These include a lack of agreement on terminology and evaluation metrics, and limitations on data and model availability. In this paper, we present a review of current research on consistency in LMs, highlight the most pressing challenges, and provide recommendations for future research. We restrict our attention to \textit{text-only} LMs, which a majority of existing research is based on. See Appendix~\ref{sec:modality} for a brief discussion on multimodal consistency.

\section{A Review of Consistency Research} 
\label{sec:research-review}


Consistency has connections to critical areas in AI research: hallucination (generating made-up information contradicting references), factuality (agreement with real-world knowledge), misinformation (false claims misleading users), and reasoning (logical coherence across statements). 
We survey literature on consistency in LMs from 2019 to 2025, focusing on peer-reviewed publications and influential preprints that explicitly address consistency metrics, theory, and enhancement.


\paragraph{Terminology}  

The terminology used to describe the consistency of LMs is often confusing, as there is not a single, commonly agreed-upon definition of consistency. Authors either come up with their own definition of the concept that aligns best with the specifics of the work they focus on, or use an overly broad definition, or sometimes just omit defining the term altogether. As a result, existing studies present multiple narrowly focused definitions of consistency that often cover very different aspects of model behavior and sometimes even contradict each other. 


Given this interest in model behavior and the implications for potential model applications, in this paper, we limit the otherwise broader concept of consistency to \textit{behavioral consistency}. 
In psychology, behavioral consistency is closely related to the predictability of 
behavior, which is equally important for the applications of LMs. Based on how behavioral consistency is approached in the literature, we categorize the different types of consistency into two large groups: logical/formal and nonlogical/informal. 


\textit{Logical consistency} in LLMs was introduced by \citet{jang2022becel}, as the ability of the model to make decisions without logical contradiction. The rules and principles of formal logic are applied to assess the behavior of a model in a methodical way, allowing for standardized and intuitive measurement. Based on these principles, \citet{jang2022becel} classified consistency into negational, symmetric, transitive, and additive types. 
Negational consistency follows the logical negation property ($p$ is true $\Leftrightarrow  \neg{p}$ is false), i.e. LM's predictions should be opposite for texts with the opposite meanings. Symmetric consistency follows the rule $f (x, y) = f (y, x)$ and implies that the predictions of an LM should be invariant to the input text swap. Transitive consistency can measure deductive reasoning ability and follows the property of transitive inference, represented as $X \rightarrow Y \land Y \rightarrow Z$ then $X \rightarrow Z$. This type of consistency was analyzed in natural language inference (NLI) tasks~\cite{li2019logic} and question-answering (Q\&A)~\cite{asai2020logic,mitchell2022enhancing}. 


\textit{Semantic consistency}, another subpart of the \citet{jang2022becel} definition, is one of the most widely used concepts in existing consistency research studies. The idea of semantic consistency is derived from the semantic equivalence property, represented as $f (X) = f (Y )$ if $X$ and $Y$ mean the same. \citet{elazar-etal-2021-measuring}, and multiple studies later on~\cite{raj2023measuring,ohmer2024form}, explored this as the ability of a model to make consistent decisions in semantically equivalent contexts.


\textit{Nonlogical or informal consistency} covers all the other definitions that do not follow the rules of formal logic. For example, \citet{bonagiri-etal-2024-sage-evaluating} highlight the importance of moral consistency, as the ability to preserve noncontradictory moral values across different situations \cite{arvanitis2020,Marcus1980-MARMDA}, in LLM alignment. Their approach consists of generating semantically equivalent scenarios and employing consistency checks to see if a target LLM gets the same Semantic Graph Entropy (SaGE) score while responding to these scenarios. 
\citet{jainetal2024} investigated norm inconsistency, defined as the condition in which LLMs apply different norms in similar situations, on applying LLMs in high-risk domains.

\textit{Informational and/or factual consistency} is another subpart of the \citet{jang2022becel} definition frequently used in consistency research. \citet{manakul2023selfcheckgptzeroresourceblackboxhallucination}
used the term \textit{informational consistency}, without explaining or defining it further, to develop a method for fact-checking the responses of black-box models. The term \textit{factual consistency} is often used in the context of automatic summarization~\cite{wang-etal-2020-asking}. Factual inconsistency is often referred to as hallucinations and/or faithfulness, i.e., models that generate new information that contradicts the source document \cite{tam-etal-2023-evaluating,maynez-etal-2020-faithfulness}. Definitions of factual consistency are often not clearly specified, and instead are replaced with human annotations.

 
A recent study \cite{parcalabescu-frank-2024-measuring} on natural language explanations contrasts faithfulness and \textit{self-consistency}. Self-consistency examines whether similar inputs produce consistent explanations---essentially measuring explanation {\it stability} across input variations. Faithfulness, meanwhile, evaluates whether the explanation behind a certain model-generated answer {\it accurately} reflects the model's reasoning process to come up with that answer. While related, they involve different evaluation approaches. Self-consistency requires testing multiple input variations (which may not generalize well across datasets) and does not necessarily involve checking for accuracy of the explanations. On the other hand, faithfulness focuses on the accuracy of individual explanations without such constraints.


\paragraph{Analyzed Tasks} 
A slim majority of studies on LM consistency investigate well-established NLP tasks. Most commonly analyzed tasks include Q\&A~\cite{mundler2024selfcontradictoryhallucinationslargelanguage, raj2023measuring, berglund2024the, Li2023BenchmarkingAI, wang-etal-2020-asking, asai-hajishirzi-2020-logic}, summarization \cite{west2024the, cui2024dcrconsistencydivideconquerreasoningconsistencyevaluation, tam-etal-2023-evaluating, wang-etal-2020-asking}, NLI \cite{jang2022becel, jang-lukasiewicz-2023-consistency, camburu-etal-2020-make, west2024the, dziri-etal-2019-evaluating} and reasoning \cite{zhang-etal-2024-self-contrast, liu2024selfcontradictoryreasoningevaluationdetection, chen2024two, wang2023selfconsistency} 
Approximately a third of existing studies do not rely on standard NLP tasks, usually using custom tasks such as generating continuations of sentences from Wikipedia \cite{mundler2024selfcontradictoryhallucinationslargelanguage}. A small number of studies employ use-case specific approaches, for example, measuring stock price prediction accuracy based on textual information such as earnings calls and news articles \cite{yang-etal-2023-measuring}. 

\paragraph{Dataset Size and Availability} 
The number of testing samples varies substantially across different studies, from a few hundred to tens of thousands. One standard approach to creating a test dataset for measuring consistency is to multiply the prompts in one or more existing benchmarks using perturbation rules or prompt templates~\citep{jang2022becel,fierro-sogaard-2022-factual}. Another approach is to enhance existing benchmarks with human- or LLM-generated annotations~\citep{liu2023score}. To do this, a common method is to create paraphrases of an existing dataset using automatic parahrasing methods~\citep{bonagiri-etal-2024-sage-evaluating} and/or human annotators~\citep{elazar-etal-2021-measuring}. The majority of testing datasets are shared publicly, although in some cases the authors only describe the dataset creation process without providing access to the actual dataset.


\paragraph{Evaluated Models} 
More than two-thirds of the studies we examined use transformer-based generative LMs with decoder-only or encoder-decoder architectures, such as the GPT and OPT series models, BART, and T5 \cite{jang2022becel, Li2023BenchmarkingAI, jang-lukasiewicz-2023-consistency, west2024the, berglund2024the, raj2023measuring, manakul2023selfcheckgptzeroresourceblackboxhallucination, mundler2024selfcontradictoryhallucinationslargelanguage, zhang-etal-2024-self-contrast, liu2024selfcontradictoryreasoningevaluationdetection, zhang-etal-2024-inconsistent, cheng2024relic, cui2024dcrconsistencydivideconquerreasoningconsistencyevaluation, tam-etal-2023-evaluating, wang2023selfconsistency, chen2024two, wang-etal-2020-asking, nie-etal-2021-like}. The parameter sizes for the models tested range from a few billion to hundreds of billion. Slightly more than half of the papers test proprietary models such as GPT-4 \citep{Li2023BenchmarkingAI, jang-lukasiewicz-2023-consistency, west2024the, berglund2024the,mundler2024selfcontradictoryhallucinationslargelanguage, zhang-etal-2024-self-contrast, liu2024selfcontradictoryreasoningevaluationdetection, zhang-etal-2024-inconsistent, cui2024dcrconsistencydivideconquerreasoningconsistencyevaluation, chen2024two}, whose exact sizes have not been publicly disclosed but in some cases are rumored to exceed a trillion parameters. 
Some studies also consider other types of LMs: about a quarter of papers \cite{jang2022becel, elazar-etal-2021-measuring, asai-hajishirzi-2020-logic, yang-etal-2023-measuring, nie-etal-2021-like, qin-etal-2021-dont} focus on encoder-only, BERT-style models such as BERT, RoBERTa, and ALBERT. 


\paragraph{Evaluation of Consistency} 

Consistency evaluation typically uses two approaches: (1) \textit{input-based sampling}, creating paraphrases or equivalent prompts to test consistent responses to similar inputs, or (2) \textit{output-based sampling}, generating multiple outputs from identical inputs. Output-based sampling with high temperature may artificially inflate inconsistency by forcing models to sample normally-avoided tokens, potentially misrepresenting model behavior.

Metrics to measure different notions of consistency typically depend on pairwise similarity metrics. They compute base metrics such as BERTScore, ROUGE, Entailment, or Contradiction for pairs of outputs given similar inputs and/or context, and aggregate over multiple pairs. In earlier studies, the base metrics were based on token-matching similarities \citep{elazar-etal-2021-measuring}. Later papers graduated to notions of semantic similarity that are robust to 
syntactic variations that can change the wording or 
structure of a phrase of text while keeping the meaning the same or similar \citep{raj2023measuring,rabinovich-etal-2023-predicting,manakul2023selfcheckgptzeroresourceblackboxhallucination}. Aggregation of a metric across pairs is typically done by simple averaging, with the exception of \citet{mundler2024selfcontradictoryhallucinationslargelanguage}, which uses sequential aggregation of contradiction scores to measure factual consistency, and \citet{raj2025improvingconsistencylargelanguage,kuhn2023semanticuncertaintylinguisticinvariances} who use semantic entropy across the entire set of outputs.



\paragraph{Challenges}
Two important aspects of consistency remain underresearched. First, current work tends to focus excessively on consistency in generations at decoding time. In this process, it ignores encoder-only models and how (in)consistent inputs shape the performance on downstream standard NLP tasks like sentiment prediction. Another underexplored direction is adversarial attacks to degrade consistency. Despite extensive research on adversarial robustness (e.g. the AdvGLUE benchmark~\citep{wang2022adversarialgluemultitaskbenchmark}) and jailbreaks, very few studies explore how inconspicuous or subtle manipulation of prompts can lead to inconsistent LLM responses~\citep{lin2024sales}. We do not yet fully understand how much malicious perturbations coupled with slightly different input text can degrade output quality.


The availability of model weights and training datasets---allowing stronger transparency and reproducibility---aid in investigating the root causes of inconsistency. 
\citet{lin2024towards} showed that analyzing the internal state of the model can improve the transparency of the model and lay the foundation for mitigating hallucinations and inconsistencies. Not only closed-weight models, but also unpublished source code and datasets make it nearly impossible to reproduce and verify claims and findings of some existing publications~\citep{semmelrock2023reproducibility}. 


\section{Discussion and Recommendations}
\label{sec:discussion}


As mentioned earlier, we need standardization of terms and definitions
for a better understanding of the progress of consistent language model development. Beyond this, we recommend the following focus areas for future research.

\vspace{-1em}
\paragraph{Multilingual Consistency}

Similarly to other topics in NLP research, the overwhelming majority of studies on LM consistency are English-based, significantly limiting our understanding of the topic. To broaden this understanding, more research is needed on both monolingual consistency in non-English languages and on cross-language consistency behaviors.




There is a substantial gap between the amount of training data available for English and that available for all other languages~\cite{ustun2024aya}. While more than 7,000 languages are spoken around the world today, an astounding 73\% of the popular datasets used to train LLMs are primarily or entirely English~\cite{longpre2023data}. This severe sampling bias in dataset construction results in disparities in model performance between languages, even in well-studied tasks~\cite{lai2023chatgpt}. Inherent differences between languages may also significantly influence the consistency of the LMs trained on them. Structural features such as word order or inflectional morphology can vary in their stability across languages~\cite{dediu2013some}. These differences can make it more difficult to train models to produce consistent output for certain languages, even when all languages are equally represented in the training data. More research is necessary to understand the effect of linguistic differences and limitations of multilingual training data on consistency in non-English languages.

Recent work has demonstrated significant challenges in \textit{cross-lingual consistency}, i.e., whether a model produces compatible or equivalent outputs when the same query is presented in different languages. \citet{shen-etal-2024-language} found that LLMs exhibit inconsistent safety behaviors across languages, with safety guardrails being more easily circumvented in non-English languages. \citet{Xing2024EvaluatingKC} observed that LLMs produce inconsistent factual information when asked about the same knowledge in different languages, suggesting knowledge representation gaps across languages. \citet{Qi2023CrossLingualCO} examined factual consistency across languages and found that languages more dissimilar to English are less likely to reflect synthetically inserted factual associations through model editing. \citet{Jin2023BetterTA} evaluated cross-lingual inconsistency specifically in healthcare questions and found discrepancies in medical advice across languages. \citet{Zhou-political-biases} explored how political biases manifest inconsistently in bilingual models, revealing that models may express different political positions depending on the input language.

These findings collectively highlight a critical gap in current LLM capabilities: the ability to maintain consistent factual information, safety guardrails, and reasoning in different languages. Cross-lingual consistency represents an important direction for future research, especially as LLMs are deployed globally across linguistic boundaries.





\paragraph{Consistency Evaluation}
Evaluating consistency has several unique challenges. Most previous studies have used automatic metrics alone to assess consistency in LMs. Although automatic evaluation can ensure objectivity and fast assessment, human evaluation is important to establish an acceptable baseline, especially in highly sensitive or subjective culture-specific applications (e.g. social appropriateness), or when automatic metrics are sufficiently high. Automatic metrics often struggle to capture the different nuances of consistency (factual, logical, semantic), while human evaluation suffers from subjectivity and cognitive biases. The contextual nature of consistency requires evaluation across multiple responses, different phrasings, and various contexts, making comprehensive assessment computationally expensive and logistically challenging. Further complicating matters, consistency evaluation interacts with other dimensions such as factuality, helpfulness, and safety---a model may be internally consistent but factually incorrect, or it may sacrifice consistency to maintain safety.

While several consistency benchmarks have recently emerged, there remains a need for more comprehensive evaluation frameworks that measure all different aspects of consistency in LMs across diverse tasks. Recent benchmarks have made important contributions but typically focus on specific consistency types or limited task domains~\citep{jang2022becel,bonagiri-etal-2024-sage-evaluating,cui2024dcrconsistencydivideconquerreasoningconsistencyevaluation,liu2024selfcontradictoryreasoningevaluationdetection,paleka2024consistencyChecksForecastersLanguage,liu2024aligninglogicmeasuringevaluating,gilhuly2025consistencyevaluationnewsarticle}. Future work should focus on developing more holistic benchmarks that address the breadth of consistency challenges outlined above.

\vspace{-.5em}
\paragraph{Impact}


Inconsistent output can cause users of language-based systems to receive conflicting or incorrect information. This is problematic in scenarios where factual accuracy is crucial \citep{tam-etal-2023-evaluating,wang2024factualitylargelanguagemodels}---such as in medical, legal, or financial contexts---especially when such information is used for decision making. In critical systems, such as autonomous vehicles or medical diagnosis support, inconsistent responses can lead to critical safety risks. In less critical applications, inconsistent responses lead to poor user experience, cause frustration, and reduce overall utility~\citep{lazar2023frustration, van2024aidenials, zhang-etal-2024-inconsistent}. Inconsistency can also reflect and magnify the underlying societal biases and stereotypes in the training data, leading to potentially discriminatory outcomes for certain user groups, amplifying unfair use and causing representational harm~\citep{blodgett2020language}. 

Inconsistency may have some advantages in specific situations. Lower degrees of consistency can lead to diverse and creative outputs, which can be valuable in tasks requiring originality or brainstorming. Inconsistency might reflect the ability of a model to adapt to different contexts or user needs, potentially providing more personalized responses. Inconsistent outputs of a model prompt users to engage more critically with generated content, avoid overreliance, and seek additional verification. This can be beneficial in educational applications, provided the level of possible inconsistency is carefully calibrated. 

\vspace{-.5em}
\paragraph{Improving Consistency}

There are surprisingly few approaches that actually increase the consistency of LMs. Current proposals to do so fall into two narrow categories. The first approach employs fine-tuning to improve consistency between multiple generations from a LM when supplied with the same or similar inputs. \citet{elazar-etal-2021-measuring} used a custom loss function, \citet{raj2025improvingconsistencylargelanguage} used knowledge distillation from more consistent teacher models, and \citet{raj2025improvingconsistencylargelanguage,zhao2024improvingrobustnesslargelanguage} used synthetic datasets of groups of consistent input-outputs. The second approach attempts to improve self-consistency, i.e. consistency between a model's reasoning process and the final answer \cite{deng2023implicitchainthoughtreasoning,wang2023selfconsistency,wei2022chain}. 

Albeit promising, the above methods primarily address symptoms rather than the fundamental causes of inconsistency. There remains a critical need for research investigating the structural basis of consistency in LMs' representational spaces, consistency-oriented pre-training, and architectures designed to maintain consistency across diverse contexts. Such foundational approaches may eliminate the trade-offs between consistency and other valuable properties like creativity and diversity.

\section{Call to Action}
\label{sec:conclusions}


We call on the research community to address several key challenges: (1) developing standardized definitions and taxonomies of consistency types; (2) creating comprehensive, multilingual, and cross-lingual benchmarks for consistency evaluation; (3) establishing robust evaluation protocols that combine automatic metrics with human evaluation; (4) investigating the relationship between consistency and other important properties such as factuality, safety, and helpfulness; and (5) developing efficient methods to enhance consistency without sacrificing other beneficial model capabilities. To this end, we emphasize the need for interdisciplinary collaboration, bringing together perspectives from linguistics, psychology, philosophy, and ethics to better understand the multifaceted nature of consistency in human and machine language use. By addressing these challenges collectively, we can move toward LMs that exhibit more reliable, trustworthy, and human-aligned behavior across diverse contexts and applications.


\bibliography{consistency-references}

\begin{thebibliography}{64}
\providecommand{\natexlab}[1]{#1}
\providecommand{\url}[1]{\texttt{#1}}
\expandafter\ifx\csname urlstyle\endcsname\relax
  \providecommand{\doi}[1]{doi: #1}\else
  \providecommand{\doi}{doi: \begingroup \urlstyle{rm}\Url}\fi

\bibitem[Arvanitis \& Kalliris(2020)Arvanitis and Kalliris]{arvanitis2020}
Arvanitis, A. and Kalliris, K.
\newblock Consistency and moral integrity: A self-determination theory perspective.
\newblock \emph{Journal of Moral Education}, 49\penalty0 (3):\penalty0 316--329, 2020.

\bibitem[Asai \& Hajishirzi(2020{\natexlab{a}})Asai and Hajishirzi]{asai-hajishirzi-2020-logic}
Asai, A. and Hajishirzi, H.
\newblock Logic-guided data augmentation and regularization for consistent question answering.
\newblock In Jurafsky, D., Chai, J., Schluter, N., and Tetreault, J. (eds.), \emph{Proceedings of the 58th Annual Meeting of the Association for Computational Linguistics}, pp.\  5642--5650, Online, July 2020{\natexlab{a}}. Association for Computational Linguistics.
\newblock \doi{10.18653/v1/2020.acl-main.499}.
\newblock URL \url{https://aclanthology.org/2020.acl-main.499}.

\bibitem[Asai \& Hajishirzi(2020{\natexlab{b}})Asai and Hajishirzi]{asai2020logic}
Asai, A. and Hajishirzi, H.
\newblock Logic-guided data augmentation and regularization for consistent question answering.
\newblock \emph{arXiv preprint arXiv:2004.10157}, 2020{\natexlab{b}}.

\bibitem[Berglund et~al.(2024)Berglund, Tong, Kaufmann, Balesni, Stickland, Korbak, and Evans]{berglund2024the}
Berglund, L., Tong, M., Kaufmann, M., Balesni, M., Stickland, A.~C., Korbak, T., and Evans, O.
\newblock The reversal curse: {LLM}s trained on {\textquotedblleft}{A} is {B}{\textquotedblright} fail to learn {\textquotedblleft}{B} is {A}{\textquotedblright}.
\newblock In \emph{The Twelfth International Conference on Learning Representations}, 2024.
\newblock URL \url{https://openreview.net/forum?id=GPKTIktA0k}.

\bibitem[Blodgett et~al.(2020)Blodgett, Barocas, Daum{\'e}~III, and Wallach]{blodgett2020language}
Blodgett, S.~L., Barocas, S., Daum{\'e}~III, H., and Wallach, H.
\newblock Language (technology) is power: A critical survey of “bias” in nlp.
\newblock In \emph{Proceedings of the 58th Annual Meeting of the Association for Computational Linguistics}, pp.\  5454--5476, 2020.

\bibitem[Bonagiri et~al.(2024)Bonagiri, Vennam, Govil, Kumaraguru, and Gaur]{bonagiri-etal-2024-sage-evaluating}
Bonagiri, V.~K., Vennam, S., Govil, P., Kumaraguru, P., and Gaur, M.
\newblock {S}a{GE}: Evaluating moral consistency in large language models.
\newblock In Calzolari, N., Kan, M.-Y., Hoste, V., Lenci, A., Sakti, S., and Xue, N. (eds.), \emph{Proceedings of the 2024 Joint International Conference on Computational Linguistics, Language Resources and Evaluation (LREC-COLING 2024)}, pp.\  14272--14284, Torino, Italia, May 2024. ELRA and ICCL.
\newblock URL \url{https://aclanthology.org/2024.lrec-main.1243}.

\bibitem[Camburu et~al.(2020)Camburu, Shillingford, Minervini, Lukasiewicz, and Blunsom]{camburu-etal-2020-make}
Camburu, O.-M., Shillingford, B., Minervini, P., Lukasiewicz, T., and Blunsom, P.
\newblock Make up your mind! adversarial generation of inconsistent natural language explanations.
\newblock In Jurafsky, D., Chai, J., Schluter, N., and Tetreault, J. (eds.), \emph{Proceedings of the 58th Annual Meeting of the Association for Computational Linguistics}, pp.\  4157--4165, Online, July 2020. Association for Computational Linguistics.
\newblock \doi{10.18653/v1/2020.acl-main.382}.
\newblock URL \url{https://aclanthology.org/2020.acl-main.382}.

\bibitem[Chen et~al.(2024)Chen, Phang, Parrish, Padmakumar, Zhao, Bowman, and Cho]{chen2024two}
Chen, A., Phang, J., Parrish, A., Padmakumar, V., Zhao, C., Bowman, S.~R., and Cho, K.
\newblock Two failures of self-consistency in the multi-step reasoning of {LLM}s.
\newblock \emph{Transactions on Machine Learning Research}, 2024.
\newblock ISSN 2835-8856.
\newblock URL \url{https://openreview.net/forum?id=5nBqY1y96B}.

\bibitem[Cheng et~al.(2024)Cheng, Zouhar, Arora, Sachan, Strobelt, and El-Assady]{cheng2024relic}
Cheng, F., Zouhar, V., Arora, S., Sachan, M., Strobelt, H., and El-Assady, M.
\newblock Relic: Investigating large language model responses using self-consistency.
\newblock In \emph{Proceedings of the CHI Conference on Human Factors in Computing Systems}, CHI '24. Association for Computing Machinery, 2024.
\newblock \doi{10.1145/3613904.3641904}.

\bibitem[Cui et~al.(2024)Cui, Zhang, Li, Damien, Das, Malin, and Kumar]{cui2024dcrconsistencydivideconquerreasoningconsistencyevaluation}
Cui, W., Zhang, J., Li, Z., Damien, L., Das, K., Malin, B., and Kumar, S.
\newblock Dcr-consistency: Divide-conquer-reasoning for consistency evaluation and improvement of large language models, 2024.
\newblock URL \url{https://arxiv.org/abs/2401.02132}.

\bibitem[Dediu \& Cysouw(2013)Dediu and Cysouw]{dediu2013some}
Dediu, D. and Cysouw, M.
\newblock Some structural aspects of language are more stable than others: A comparison of seven methods.
\newblock \emph{PloS one}, 8\penalty0 (1):\penalty0 e55009, 2013.

\bibitem[Deng et~al.(2023)Deng, Prasad, Fernandez, Smolensky, Chaudhary, and Shieber]{deng2023implicitchainthoughtreasoning}
Deng, Y., Prasad, K., Fernandez, R., Smolensky, P., Chaudhary, V., and Shieber, S.
\newblock Implicit chain of thought reasoning via knowledge distillation, 2023.
\newblock URL \url{https://arxiv.org/abs/2311.01460}.

\bibitem[Dziri et~al.(2019)Dziri, Kamalloo, Mathewson, and Zaiane]{dziri-etal-2019-evaluating}
Dziri, N., Kamalloo, E., Mathewson, K., and Zaiane, O.
\newblock Evaluating coherence in dialogue systems using entailment.
\newblock In Burstein, J., Doran, C., and Solorio, T. (eds.), \emph{Proceedings of the 2019 Conference of the North {A}merican Chapter of the Association for Computational Linguistics: Human Language Technologies, Volume 1 (Long and Short Papers)}, pp.\  3806--3812, Minneapolis, Minnesota, June 2019. Association for Computational Linguistics.
\newblock \doi{10.18653/v1/N19-1381}.
\newblock URL \url{https://aclanthology.org/N19-1381}.

\bibitem[Elazar et~al.(2021)Elazar, Kassner, Ravfogel, Ravichander, Hovy, Sch{\"u}tze, and Goldberg]{elazar-etal-2021-measuring}
Elazar, Y., Kassner, N., Ravfogel, S., Ravichander, A., Hovy, E., Sch{\"u}tze, H., and Goldberg, Y.
\newblock Measuring and improving consistency in pretrained language models.
\newblock \emph{Transactions of the Association for Computational Linguistics}, 9:\penalty0 1012--1031, 2021.
\newblock \doi{10.1162/tacl_a_00410}.
\newblock URL \url{https://aclanthology.org/2021.tacl-1.60}.

\bibitem[Fierro \& S{\o}gaard(2022)Fierro and S{\o}gaard]{fierro-sogaard-2022-factual}
Fierro, C. and S{\o}gaard, A.
\newblock Factual consistency of multilingual pretrained language models.
\newblock In \emph{Findings of the Association for Computational Linguistics: ACL 2022}, pp.\  3046--3052, Dublin, Ireland, May 2022. Association for Computational Linguistics.
\newblock URL \url{https://aclanthology.org/2022.findings-acl.240}.

\bibitem[Gilhuly \& Shahzad(2025)Gilhuly and Shahzad]{gilhuly2025consistencyevaluationnewsarticle}
Gilhuly, C. and Shahzad, H.
\newblock Consistency evaluation of news article summaries generated by large (and small) language models.
\newblock \emph{arXiv preprint arXiv:2502.20647}, 2025.

\bibitem[Jain et~al.(2025)Jain, Calacci, and Wilson]{jainetal2024}
Jain, S., Calacci, D., and Wilson, A.
\newblock \emph{As an AI Language Model, "Yes I Would Recommend Calling the Police": Norm Inconsistency in LLM Decision-Making}, pp.\  624–633.
\newblock AAAI Press, 2025.

\bibitem[Jang \& Lukasiewicz(2023)Jang and Lukasiewicz]{jang-lukasiewicz-2023-consistency}
Jang, M. and Lukasiewicz, T.
\newblock Consistency analysis of {C}hat{GPT}.
\newblock In Bouamor, H., Pino, J., and Bali, K. (eds.), \emph{Proceedings of the 2023 Conference on Empirical Methods in Natural Language Processing}, pp.\  15970--15985, Singapore, December 2023. Association for Computational Linguistics.
\newblock \doi{10.18653/v1/2023.emnlp-main.991}.
\newblock URL \url{https://aclanthology.org/2023.emnlp-main.991}.

\bibitem[Jang et~al.(2022)Jang, Kwon, and Lukasiewicz]{jang2022becel}
Jang, M., Kwon, D.~S., and Lukasiewicz, T.
\newblock Becel: Benchmark for consistency evaluation of language models.
\newblock In \emph{Proceedings of the 29th International Conference on Computational Linguistics}, pp.\  3680--3696, 2022.

\bibitem[Jin et~al.(2023)Jin, Chandra, Verma, Hu, Choudhury, and Kumar]{Jin2023BetterTA}
Jin, Y., Chandra, M., Verma, G., Hu, Y., Choudhury, M.~D., and Kumar, S.
\newblock Better to ask in english: Cross-lingual evaluation of large language models for healthcare queries.
\newblock \emph{Proceedings of the ACM on Web Conference 2024}, 2023.
\newblock URL \url{https://api.semanticscholar.org/CorpusID:264405758}.

\bibitem[Kim et~al.(2025)Kim, Vaughan, Liao, Lombrozo, and Russakovsky]{kim2025fostering}
Kim, S. S.~Y., Vaughan, J.~W., Liao, Q.~V., Lombrozo, T., and Russakovsky, O.
\newblock Fostering appropriate reliance on large language models: The role of explanations, sources, and inconsistencies.
\newblock In \emph{CHI Conference on Human Factors in Computing Systems}, pp.\  1--26, Yokohama, Japan, 2025. ACM.
\newblock ISBN 979-8-4007-1394-1/25/04.
\newblock \doi{10.1145/3706598.3714020}.

\bibitem[Kuhn et~al.(2023)Kuhn, Gal, and Farquhar]{kuhn2023semanticuncertaintylinguisticinvariances}
Kuhn, L., Gal, Y., and Farquhar, S.
\newblock Semantic uncertainty: Linguistic invariances for uncertainty estimation in natural language generation, 2023.
\newblock URL \url{https://arxiv.org/abs/2302.09664}.

\bibitem[Lai et~al.(2023)Lai, Ngo, Veyseh, Man, Dernoncourt, Bui, and Nguyen]{lai2023chatgpt}
Lai, V., Ngo, N., Veyseh, A. P.~B., Man, H., Dernoncourt, F., Bui, T., and Nguyen, T.
\newblock Chatgpt beyond english: Towards a comprehensive evaluation of large language models in multilingual learning.
\newblock In \emph{Findings of the Association for Computational Linguistics: EMNLP 2023}, pp.\  13171--13189, 2023.

\bibitem[Lazar et~al.(2023)Lazar, Feng, Lazar, and Wentz]{lazar2023frustration}
Lazar, J., Feng, J.~H., Lazar, A., and Wentz, B.
\newblock Frustration: Still a common user experience.
\newblock \emph{ACM Transactions on Computer-Human Interaction}, 30\penalty0 (3):\penalty0 1--22, 2023.
\newblock \doi{10.1145/3582432}.

\bibitem[Li et~al.(2019)Li, Gupta, Mehta, and Srikumar]{li2019logic}
Li, T., Gupta, V., Mehta, M., and Srikumar, V.
\newblock A logic-driven framework for consistency of neural models.
\newblock \emph{arXiv preprint arXiv:1909.00126}, 2019.

\bibitem[Li et~al.(2023)Li, Shrivastava, Li, Hashimoto, and Liang]{Li2023BenchmarkingAI}
Li, X.~L., Shrivastava, V., Li, S., Hashimoto, T., and Liang, P.
\newblock Benchmarking and improving generator-validator consistency of language models.
\newblock \emph{ArXiv}, abs/2310.01846, 2023.
\newblock URL \url{https://api.semanticscholar.org/CorpusID:263609159}.

\bibitem[Lin et~al.(2024{\natexlab{a}})Lin, Gerchanovsky, Akgul, Bauer, Fredrikson, and Wang]{lin2024sales}
Lin, W., Gerchanovsky, A., Akgul, O., Bauer, L., Fredrikson, M., and Wang, Z.
\newblock Llm whisperer: An inconspicuous attack to bias llm responses.
\newblock \emph{arXiv preprint arXiv:2406.04755}, 2024{\natexlab{a}}.

\bibitem[Lin et~al.(2024{\natexlab{b}})Lin, Guan, Zhang, Zhang, Li, and Zhang]{lin2024towards}
Lin, Z., Guan, S., Zhang, W., Zhang, H., Li, Y., and Zhang, H.
\newblock Towards trustworthy llms: a review on debiasing and dehallucinating in large language models.
\newblock \emph{Artificial Intelligence Review}, 57\penalty0 (9):\penalty0 1--50, 2024{\natexlab{b}}.

\bibitem[Liu et~al.(2024{\natexlab{a}})Liu, Li, Zhang, Fan, Zhou, and Liang]{liu2024aligninglogicmeasuringevaluating}
Liu, Y., Li, Y., Zhang, J., Fan, P., Zhou, Y., and Liang, K.
\newblock Aligning with logic: Measuring, evaluating and improving logical preference consistency in large language models.
\newblock \emph{arXiv preprint arXiv:2410.02205}, 2024{\natexlab{a}}.
\newblock Updated 2025.

\bibitem[Liu et~al.(2023)Liu, Lee, Du, Sanyal, and Zhao]{liu2023score}
Liu, Z., Lee, I., Du, Y., Sanyal, S., and Zhao, J.
\newblock Score: A framework for self-contradictory reasoning evaluation.
\newblock \emph{arXiv preprint arXiv:2311.09603}, 2023.

\bibitem[Liu et~al.(2024{\natexlab{b}})Liu, Lee, Du, Sanyal, and Zhao]{liu2024selfcontradictoryreasoningevaluationdetection}
Liu, Z., Lee, I., Du, Y., Sanyal, S., and Zhao, J.
\newblock Self-contradictory reasoning evaluation and detection, 2024{\natexlab{b}}.
\newblock URL \url{https://arxiv.org/abs/2311.09603}.

\bibitem[Longpre et~al.(2023)Longpre, Mahari, Chen, Obeng-Marnu, Sileo, Brannon, Muennighoff, Khazam, Kabbara, Perisetla, et~al.]{longpre2023data}
Longpre, S., Mahari, R., Chen, A., Obeng-Marnu, N., Sileo, D., Brannon, W., Muennighoff, N., Khazam, N., Kabbara, J., Perisetla, K., et~al.
\newblock The data provenance initiative: A large scale audit of dataset licensing \& attribution in ai.
\newblock \emph{arXiv preprint arXiv:2310.16787}, 2023.

\bibitem[Manakul et~al.(2023)Manakul, Liusie, and Gales]{manakul2023selfcheckgptzeroresourceblackboxhallucination}
Manakul, P., Liusie, A., and Gales, M. J.~F.
\newblock Selfcheckgpt: Zero-resource black-box hallucination detection for generative large language models, 2023.
\newblock URL \url{https://arxiv.org/abs/2303.08896}.

\bibitem[Marcus(1980)]{Marcus1980-MARMDA}
Marcus, R.~B.
\newblock Moral dilemmas and consistency.
\newblock \emph{Journal of Philosophy}, 77\penalty0 (3):\penalty0 121--136, 1980.
\newblock \doi{10.2307/2025665}.

\bibitem[Maynez et~al.(2020)Maynez, Narayan, Bohnet, and McDonald]{maynez-etal-2020-faithfulness}
Maynez, J., Narayan, S., Bohnet, B., and McDonald, R.
\newblock On faithfulness and factuality in abstractive summarization.
\newblock In Jurafsky, D., Chai, J., Schluter, N., and Tetreault, J. (eds.), \emph{Proceedings of the 58th Annual Meeting of the Association for Computational Linguistics}, pp.\  1906--1919, Online, July 2020. Association for Computational Linguistics.
\newblock \doi{10.18653/v1/2020.acl-main.173}.
\newblock URL \url{https://aclanthology.org/2020.acl-main.173/}.

\bibitem[Mitchell et~al.(2022)Mitchell, Noh, Li, Armstrong, Agarwal, Liu, Finn, and Manning]{mitchell2022enhancing}
Mitchell, E., Noh, J.~J., Li, S., Armstrong, W.~S., Agarwal, A., Liu, P., Finn, C., and Manning, C.~D.
\newblock Enhancing self-consistency and performance of pre-trained language models through natural language inference.
\newblock \emph{arXiv preprint arXiv:2211.11875}, 2022.

\bibitem[Mündler et~al.(2024)Mündler, He, Jenko, and Vechev]{mundler2024selfcontradictoryhallucinationslargelanguage}
Mündler, N., He, J., Jenko, S., and Vechev, M.
\newblock Self-contradictory hallucinations of large language models: Evaluation, detection and mitigation, 2024.
\newblock URL \url{https://arxiv.org/abs/2305.15852}.

\bibitem[Nie et~al.(2021)Nie, Williamson, Bansal, Kiela, and Weston]{nie-etal-2021-like}
Nie, Y., Williamson, M., Bansal, M., Kiela, D., and Weston, J.
\newblock {I} like fish, especially dolphins: Addressing contradictions in dialogue modeling.
\newblock In Zong, C., Xia, F., Li, W., and Navigli, R. (eds.), \emph{Proceedings of the 59th Annual Meeting of the Association for Computational Linguistics and the 11th International Joint Conference on Natural Language Processing (Volume 1: Long Papers)}, pp.\  1699--1713, Online, August 2021. Association for Computational Linguistics.
\newblock \doi{10.18653/v1/2021.acl-long.134}.
\newblock URL \url{https://aclanthology.org/2021.acl-long.134}.

\bibitem[Ohmer et~al.(2024)Ohmer, Bruni, and Hupkes]{ohmer2024form}
Ohmer, X., Bruni, E., and Hupkes, D.
\newblock From form (s) to meaning: Probing the semantic depths of language models using multisense consistency.
\newblock \emph{Computational Linguistics}, pp.\  1--51, 2024.

\bibitem[Paleka et~al.(2024)Paleka, Hadjikyriacou, Daneshjou, Gleave, and Steinhardt]{paleka2024consistencyChecksForecastersLanguage}
Paleka, D., Hadjikyriacou, A., Daneshjou, R., Gleave, A., and Steinhardt, J.
\newblock Consistency checks for language model forecasters.
\newblock \emph{arXiv preprint arXiv:2412.18544}, 2024.
\newblock Updated January 2025.

\bibitem[Parcalabescu \& Frank(2024)Parcalabescu and Frank]{parcalabescu-frank-2024-measuring}
Parcalabescu, L. and Frank, A.
\newblock On measuring faithfulness or self-consistency of natural language explanations.
\newblock In Ku, L.-W., Martins, A., and Srikumar, V. (eds.), \emph{Proceedings of the 62nd Annual Meeting of the Association for Computational Linguistics (Volume 1: Long Papers)}, pp.\  6048--6089, Bangkok, Thailand, August 2024. Association for Computational Linguistics.
\newblock URL \url{https://aclanthology.org/2024.acl-long.329}.

\bibitem[Qi et~al.(2023)Qi, Fern'andez, and Bisazza]{Qi2023CrossLingualCO}
Qi, J., Fern'andez, R., and Bisazza, A.
\newblock Cross-lingual consistency of factual knowledge in multilingual language models.
\newblock In \emph{Conference on Empirical Methods in Natural Language Processing}, 2023.
\newblock URL \url{https://api.semanticscholar.org/CorpusID:264145744}.

\bibitem[Qin et~al.(2021)Qin, Xie, Huang, Chen, Xu, and Che]{qin-etal-2021-dont}
Qin, L., Xie, T., Huang, S., Chen, Q., Xu, X., and Che, W.
\newblock Don{'}t be contradicted with anything! {CI}-{T}o{D}: Towards benchmarking consistency for task-oriented dialogue system.
\newblock In Moens, M.-F., Huang, X., Specia, L., and Yih, S. W.-t. (eds.), \emph{Proceedings of the 2021 Conference on Empirical Methods in Natural Language Processing}, pp.\  2357--2367, Online and Punta Cana, Dominican Republic, November 2021. Association for Computational Linguistics.
\newblock \doi{10.18653/v1/2021.emnlp-main.182}.
\newblock URL \url{https://aclanthology.org/2021.emnlp-main.182}.

\bibitem[Rabinovich et~al.(2023)Rabinovich, Ackerman, Raz, Farchi, and Anaby~Tavor]{rabinovich-etal-2023-predicting}
Rabinovich, E., Ackerman, S., Raz, O., Farchi, E., and Anaby~Tavor, A.
\newblock Predicting question-answering performance of large language models through semantic consistency.
\newblock In Gehrmann, S., Wang, A., Sedoc, J., Clark, E., Dhole, K., Chandu, K.~R., Santus, E., and Sedghamiz, H. (eds.), \emph{Proceedings of the Third Workshop on Natural Language Generation, Evaluation, and Metrics (GEM)}, pp.\  138--154, Singapore, December 2023. Association for Computational Linguistics.
\newblock URL \url{https://aclanthology.org/2023.gem-1.12}.

\bibitem[Raj et~al.(2022)Raj, Rosati, and Majumdar]{raj2023measuring}
Raj, H., Rosati, D., and Majumdar, S.
\newblock Measuring reliability of large language models through semantic consistency, 2022.

\bibitem[Raj et~al.(2025)Raj, Gupta, Rosati, and Majumdar]{raj2025improvingconsistencylargelanguage}
Raj, H., Gupta, V., Rosati, D., and Majumdar, S.
\newblock Improving consistency in large language models through chain of guidance.
\newblock \emph{Transactions on Machine Learning Research}, 2025.
\newblock URL \url{https://arxiv.org/abs/2502.15924}.

\bibitem[Semmelrock et~al.(2023)Semmelrock, Kopeinik, Theiler, Ross-Hellauer, and Kowald]{semmelrock2023reproducibility}
Semmelrock, H., Kopeinik, S., Theiler, D., Ross-Hellauer, T., and Kowald, D.
\newblock Reproducibility in machine learning-driven research.
\newblock \emph{arXiv preprint arXiv:2307.10320}, 2023.

\bibitem[Shen et~al.(2024)Shen, Tan, Chen, Chen, Zhang, Xu, Zheng, Koehn, and Khashabi]{shen-etal-2024-language}
Shen, L., Tan, W., Chen, S., Chen, Y., Zhang, J., Xu, H., Zheng, B., Koehn, P., and Khashabi, D.
\newblock The language barrier: Dissecting safety challenges of {LLM}s in multilingual contexts.
\newblock In Ku, L.-W., Martins, A., and Srikumar, V. (eds.), \emph{Findings of the Association for Computational Linguistics: ACL 2024}, pp.\  2668--2680, Bangkok, Thailand, August 2024. Association for Computational Linguistics.
\newblock \doi{10.18653/v1/2024.findings-acl.156}.
\newblock URL \url{https://aclanthology.org/2024.findings-acl.156}.

\bibitem[Tam et~al.(2023)Tam, Mascarenhas, Zhang, Kwan, Bansal, and Raffel]{tam-etal-2023-evaluating}
Tam, D., Mascarenhas, A., Zhang, S., Kwan, S., Bansal, M., and Raffel, C.
\newblock Evaluating the factual consistency of large language models through news summarization.
\newblock In Rogers, A., Boyd-Graber, J., and Okazaki, N. (eds.), \emph{Findings of the Association for Computational Linguistics: ACL 2023}, pp.\  5220--5255, Toronto, Canada, July 2023. Association for Computational Linguistics.

\bibitem[Tan et~al.(2022)Tan, Yang, Ye, Wang, Yan, Nguyen, and Huang]{tan2022ssd}
Tan, Z., Yang, X., Ye, Z., Wang, Q., Yan, Y., Nguyen, A., and Huang, K.
\newblock Ssd: Towards better text-image consistency metric in text-to-image generation.
\newblock \emph{arXiv preprint arXiv:2210.15235}, 2022.

\bibitem[{\"U}st{\"u}n et~al.(2024){\"U}st{\"u}n, Aryabumi, Yong, Ko, D'souza, Onilude, Bhandari, Singh, Ooi, Kayid, et~al.]{ustun2024aya}
{\"U}st{\"u}n, A., Aryabumi, V., Yong, Z.-X., Ko, W.-Y., D'souza, D., Onilude, G., Bhandari, N., Singh, S., Ooi, H.-L., Kayid, A., et~al.
\newblock Aya model: An instruction finetuned open-access multilingual language model.
\newblock \emph{arXiv preprint arXiv:2402.07827}, 2024.

\bibitem[van Bergen et~al.(2024)van Bergen, van~der Schalk, Kökciyan, Otterbacher, Haider, and Terzimehić]{van2024aidenials}
van Bergen, R., van~der Schalk, B., Kökciyan, N., Otterbacher, J., Haider, J., and Terzimehić, N.
\newblock "as an ai language model, i cannot": Investigating llm denials of user requests.
\newblock In \emph{Proceedings of the 2024 CHI Conference on Human Factors in Computing Systems}, pp.\  1--16. ACM, 2024.
\newblock \doi{10.1145/3613904.3642135}.

\bibitem[Wang et~al.(2020)Wang, Cho, and Lewis]{wang-etal-2020-asking}
Wang, A., Cho, K., and Lewis, M.
\newblock Asking and answering questions to evaluate the factual consistency of summaries.
\newblock In Jurafsky, D., Chai, J., Schluter, N., and Tetreault, J. (eds.), \emph{Proceedings of the 58th Annual Meeting of the Association for Computational Linguistics}, pp.\  5008--5020, Online, July 2020. Association for Computational Linguistics.
\newblock \doi{10.18653/v1/2020.acl-main.450}.
\newblock URL \url{https://aclanthology.org/2020.acl-main.450}.

\bibitem[Wang et~al.(2022)Wang, Xu, Wang, Gan, Cheng, Gao, Awadallah, and Li]{wang2022adversarialgluemultitaskbenchmark}
Wang, B., Xu, C., Wang, S., Gan, Z., Cheng, Y., Gao, J., Awadallah, A.~H., and Li, B.
\newblock Adversarial glue: A multi-task benchmark for robustness evaluation of language models, 2022.
\newblock URL \url{https://arxiv.org/abs/2111.02840}.

\bibitem[Wang et~al.(2023)Wang, Wei, Schuurmans, Le, Chi, Narang, Chowdhery, and Zhou]{wang2023selfconsistency}
Wang, X., Wei, J., Schuurmans, D., Le, Q.~V., Chi, E.~H., Narang, S., Chowdhery, A., and Zhou, D.
\newblock Self-consistency improves chain of thought reasoning in language models.
\newblock In \emph{The Eleventh International Conference on Learning Representations}, 2023.
\newblock URL \url{https://openreview.net/forum?id=1PL1NIMMrw}.

\bibitem[Wang et~al.(2024)Wang, Wang, Manzoor, Liu, Georgiev, Das, and Nakov]{wang2024factualitylargelanguagemodels}
Wang, Y., Wang, M., Manzoor, M.~A., Liu, F., Georgiev, G., Das, R.~J., and Nakov, P.
\newblock Factuality of large language models in the year 2024, 2024.
\newblock URL \url{https://arxiv.org/abs/2402.02420}.

\bibitem[Wei et~al.(2022)Wei, Wang, Schuurmans, Bosma, Ichter, Xia, Chi, Le, and Zhou]{wei2022chain}
Wei, J., Wang, X., Schuurmans, D., Bosma, M., Ichter, B., Xia, F., Chi, E., Le, Q., and Zhou, D.
\newblock Chain of thought prompting elicits reasoning in large language models.
\newblock \emph{arXiv preprint arXiv:2201.11903}, 2022.

\bibitem[West et~al.(2024)West, Lu, Dziri, Brahman, Li, Hwang, Jiang, Fisher, Ravichander, Chandu, Newman, Koh, Ettinger, and Choi]{west2024the}
West, P., Lu, X., Dziri, N., Brahman, F., Li, L., Hwang, J.~D., Jiang, L., Fisher, J., Ravichander, A., Chandu, K., Newman, B., Koh, P.~W., Ettinger, A., and Choi, Y.
\newblock The generative {AI} paradox: {\textquotedblleft}what it can create, it may not understand{\textquotedblright}.
\newblock In \emph{The Twelfth International Conference on Learning Representations}, 2024.
\newblock URL \url{https://openreview.net/forum?id=CF8H8MS5P8}.

\bibitem[Xing et~al.(2024)Xing, He, Xu, Wang, Wang, and Hong]{Xing2024EvaluatingKC}
Xing, X., He, Z., Xu, H., Wang, X., Wang, R., and Hong, Y.
\newblock Evaluating knowledge-based cross-lingual inconsistency in large language models.
\newblock \emph{ArXiv}, abs/2407.01358, 2024.
\newblock URL \url{https://api.semanticscholar.org/CorpusID:270870062}.

\bibitem[Yang et~al.(2023)Yang, Ma, and Zhang]{yang-etal-2023-measuring}
Yang, L., Ma, Y., and Zhang, Y.
\newblock Measuring consistency in text-based financial forecasting models.
\newblock In Rogers, A., Boyd-Graber, J., and Okazaki, N. (eds.), \emph{Proceedings of the 61st Annual Meeting of the Association for Computational Linguistics (Volume 1: Long Papers)}, pp.\  13751--13765, Toronto, Canada, July 2023. Association for Computational Linguistics.
\newblock \doi{10.18653/v1/2023.acl-long.769}.
\newblock URL \url{https://aclanthology.org/2023.acl-long.769}.

\bibitem[Zhang et~al.(2024{\natexlab{a}})Zhang, Jin, Song, Mi, and Yu]{zhang-etal-2024-inconsistent}
Zhang, M., Jin, L., Song, L., Mi, H., and Yu, D.
\newblock Inconsistent dialogue responses and how to recover from them.
\newblock In Graham, Y. and Purver, M. (eds.), \emph{Findings of the Association for Computational Linguistics: EACL 2024}, pp.\  220--230, St. Julian{'}s, Malta, March 2024{\natexlab{a}}. Association for Computational Linguistics.
\newblock URL \url{https://aclanthology.org/2024.findings-eacl.16}.

\bibitem[Zhang et~al.(2024{\natexlab{b}})Zhang, Shen, Wu, Peng, Wang, Zhuang, and Lu]{zhang-etal-2024-self-contrast}
Zhang, W., Shen, Y., Wu, L., Peng, Q., Wang, J., Zhuang, Y., and Lu, W.
\newblock Self-contrast: Better reflection through inconsistent solving perspectives.
\newblock In Ku, L.-W., Martins, A., and Srikumar, V. (eds.), \emph{Proceedings of the 62nd Annual Meeting of the Association for Computational Linguistics (Volume 1: Long Papers)}, pp.\  3602--3622, Bangkok, Thailand, August 2024{\natexlab{b}}. Association for Computational Linguistics.
\newblock URL \url{https://aclanthology.org/2024.acl-long.197}.

\bibitem[Zhao et~al.(2024)Zhao, Yan, Sun, Xing, Wang, Meng, Cheng, Ren, and Yin]{zhao2024improvingrobustnesslargelanguage}
Zhao, Y., Yan, L., Sun, W., Xing, G., Wang, S., Meng, C., Cheng, Z., Ren, Z., and Yin, D.
\newblock Improving the robustness of large language models via consistency alignment, 2024.
\newblock URL \url{https://arxiv.org/abs/2403.14221}.

\bibitem[Zhou \& Zhang(2024)Zhou and Zhang]{Zhou-political-biases}
Zhou, D. and Zhang, Y.
\newblock Political biases and inconsistencies in bilingual gpt models—the cases of the u.s. and china.
\newblock \emph{Scientific Reports}, 14, 10 2024.
\newblock \doi{10.1038/s41598-024-76395-w}.

\end{thebibliography}
\bibliographystyle{icml2025}

\appendix
\section{Appendix: Multimodal Consistency}
\label{sec:modality}

Until 2022, every consistency study was analyzing robustness of LMs to various text perturbations or to semantically equivalent texts only. Starting in 2022, some interest in non-textual modalities started to appear that comes primarily from text-to-image model analysis. For example, \citet{tan2022ssd} explores the challenge of generating consistent and high-quality images from given texts in the task of visual-language understanding and highlights the need to design a better text-image consistency metric, a problem that remains under-explored in the community. In their study, \citet{tan2022ssd} present a novel CLIP-based metric named Semantic Similarity Distance (SSD) that leads to significantly better text-image consistency while maintaining decent image quality. The attempts to quantify the consistency in text-to-image models are continued by \citet{berglund2024the}, which proposes a novel semantic consistency score for image generation that has strong agreement with human annotators. Recently, there was an attempt to evaluate the understanding capability of generative models in both language and vision domains~\cite{west2024the}. \citet{west2024the} conducted interrogative evaluation of image understanding models via visual question answering in an open-ended setting. They investigated whether the models produce consistent output when interrogated about the content of their generated image, and figured out that although models can outperform humans in generation, they regularly show evidence of inconsistency between their generation and understanding performance.

\end{document}